\documentclass[conference]{IEEEtran}
\usepackage{blindtext, graphicx,url,subcaption}
\usepackage[justification=centering]{caption}
\usepackage{epstopdf}
\ifCLASSINFOpdf
  $\usepackage[pdftex]{graphicx}
\else
\fi

\begin{document}
%
\title{A novel and effective scoring scheme for structure classification and pairwise similarity measurement}

\author{\IEEEauthorblockN{Rezaul Karim\IEEEauthorrefmark{1},
Md. Momin Al Aziz\IEEEauthorrefmark{2},Swakkhar Shatabda\IEEEauthorrefmark{3} and
M. Sohel Rahman\IEEEauthorrefmark{1}}
\IEEEauthorblockA{
 \IEEEauthorrefmark{1}Department of Computer Science and Engineering, Bangladesh University of Engineering and Technology\\
\IEEEauthorrefmark{2}Department of Computer Science, University of Manitoba, Canada\\
\IEEEauthorrefmark{3}Department of Computer Science and Engineering, United International University}}


%


\maketitle

\begin{abstract}
Protein tertiary structure defines its functions, classification and binding sites. Similar structural characteristics between two proteins often lead to the similar characteristics thereof. Determining structural similarity accurately in real time is a crucial research issue. In this paper, we present a novel and effective scoring scheme that is dependent on novel features extracted from protein alpha carbon distance matrices. Our scoring scheme is inspired from pattern recognition and computer vision. Our method is significantly better than the current state of the art methods in terms of family match of pairs of protein structures and other statistical measurements. The effectiveness of our method is tested on standard benchmark structures. A web service is available at \url{http://research.buet.ac.bd:8080/Comograd/score.html} where you can get the similarity measurement score between two protein structures based on our method.
\end{abstract}


%
\IEEEpeerreviewmaketitle

\section{Introduction}
\label{Introduction}
Pairwise protein structure comparison is a prerequisite step to find structural neighbors of a protein. Protein tertiary structure comparison and structural neighbor search have enormous significance in many applications of modern structural biology, drug discovery, drug design and other fields. This is especially significant because a structure is more conserved than the protein sequence \cite{chothia1986relation}. Some of the major applications in this regard include function prediction of novel protein, topological clustering and classification of proteins \cite{greene2007cath}, binding site prediction \cite{mukherjee2009mm,gao2008dbd}, drug screening \cite{shoichet2012structure} and protein based \textit{in silico} drug design \cite{kitchen2004docking}. Due to the present volume of available known protein structures and the pace at which novel structures are being discovered with the advancement of modern crystallographic technologies (X-ray, NMR), the demand for better methods for automated structural comparison is more than ever.
\par
A lot of research works in the literature have focused on protein structure comparison during the past few decades \cite{hasegawa2009advances,biasini2014swiss}. The main goal of all these proposed methods is to find a similarity measurement (i.e., numeric score) between two protein structures. In other words, these methods define a useful scoring function with regards to the similarity or dissimilarity of the structures under comparison. Most of the scoring functions used in these methods are based on aligning two structures under comparison with some set of rules and constraints (e.g., distance constraints) or with some rotation. These methods differ in alignment rules, constraints and the functions used for calculating the score.
\par
However, a major drawback of these approaches with regards to the global structure similarity comparison is that these are much sensitive to local structure dissimilarity. Furthermore, these approaches need to find an optimal alignment before computing the score despite that often there are no remarkable alignment possible. Also finding proper alignment between two protein structures is computationally expensive if the size of the protein is large~\cite{godzik1996structural}. In spite of these inherent limitations and drawbacks, there exist a number of such methods in the literature and the most notable ones among these are \textit{DALI} \cite{holm1997dali,holm1993protein}, \textit{CE} \cite{shindyalov1998protein}, \textit{TM Align} \cite{xu2010significant,zhang2005tm} and \textit{SP Align} \cite{yang2012new}.
\par
Some interesting and appealing approaches have recently been presented in the literature applying ideas from computational geometry, graph theory, computer vision, pattern recognition and machine learning. Most of these methods do not align two structures beforehand to find the alignment score; rather they use some smart feature sets to compute a score that is representative of structural (dis)similarity. These approaches are less sensitive to local structure dissimilarity and often are computationally more efficient. Some of the well known methods in this category are \textit{MatAlign} \cite{aung2006matalign} and \textit{MASASW} \cite{6051424}.
\par
All these attempts gave birth to various scoring methods that answer how structurally (dis)similar two proteins are. However, in the literature we find only a few reports providing experimental verification of the statistical significance of these scoring functions and methods. In this paper, we introduce and analyze \emph{CoMOGPhog score}, a new scoring function based on the features presented in our prior work in \cite{Karim2015}. We believe the major potential advantage of CoMOGPhog score is its high scalability, because, unlike the alignment based approaches mentioned above, here the computation time does not depend on the protein size. 
We analyze the statistical significance of CoMOGPhog score with respect to a number of statistical measures that are widely used in statistics and machine learning. We also compare the performance of our score with the most popular methods and state of the art alignment scoring approaches. The scoring module is available as a web service at \url{http://research.buet.ac.bd:8080/Comograd/score.html}
\par
\par
The rest of the paper is organized as follows. We present an extensive review of the related works to comprehend the development in protein structure comparison in Section~\ref{secRW} . The details of our method and materials are given in Section \ref{secMM}. Results and discussion on the findings are given in Section~\ref{secRD}. Finally we briefly draw conclusion in Section \ref{secCon}.

\section{Related Works}
\label{secRW}
There exist a number of attempts in the literature to compare protein structures as three dimensional objects superimposing on one another. These methods opened up a path to address this problem by proposing a score function based on different distance metrics as a measure of the similarity or dissimilarity. Global distance score (GDT) is one of the notable classical methods to compare protein structures using superimposing and to calculate a distance score of the alignment. Some of the major contributions on this field are briefly reviewed below.

\subsection{DALI}
DALI \cite{holm1997dali}, or \textit{distance alignment matrix method} finds an optimal alignment between two structures and then calculates an alignment score. It breaks the input structure into hexapeptide fragments and calculates a distance matrix by evaluating the contact patterns between successive fragments.~\cite{holm1993protein}.

Advanced methods of similar approach finds an optimal alignment between two structure and then calculates some alignment score. A common and popular structural alignment method is the DALI \cite{holm1997dali}, or distance alignment matrix method, which breaks the input structures into hexapeptide fragments and calculates a distance matrix by evaluating the contact patterns between successive fragments.~\cite{holm1993protein}.
\subsection{CE}
CE or \textit{Combinatorial Extension} method \cite{shindyalov1998protein} is similar to DALI in that it too breaks each structure in the query set into a series of fragments that it then attempts to reassemble into a complete alignment. A series of pairwise combinations of fragments called aligned fragment pairs, or AFPs, are used to define a similarity matrix through which an optimal path is generated to identify the final alignment. Only AFPs that meet given criteria for local similarity are included in the matrix as a means of reducing the necessary search space and thereby increasing efficiency \cite{shindyalov1998protein}. However, in spite of having good accuracy it is impossible to implement these two methods as a real time web service due to their huge computational cost.
\subsection{Sequential Structure Alignment Program}
The SSAP (Sequential Structure Alignment Program) method\cite{orengo1996ssap} uses double dynamic programming to produce a structural alignment based on atom-to-atom vectors in structure space. Instead of the alpha carbons typically used in structural alignment, SSAP constructs its vectors from the beta carbons for all residues except glycine, a method which thus takes into account the rotameric state of each residue as well as its location along the backbone. SSAP works by first constructing a series of inter-residue distance vectors between each residue and its nearest non-contiguous neighbors on each protein. A series of matrices are then constructed containing the vector differences between neighbors for each pair of residues for which vectors were constructed. Dynamic programming applied to each resulting matrix determines a series of optimal local alignments which are then summed into a `summary' matrix to which dynamic programming is applied again to determine the overall structural alignment.
\subsection{FATCAT}
FATCAT \cite{ye2003flexible} or \textit{F}lexible structure \textit{A}lignmen\textit{T} by \textit{C}haining \textit{A}ligned fragment pairs with \textit{T}wists is a unique approach in a sense that it took the structural rearrangements that the proteins go through into account. This is different as in most methods the structure is treated like a fixed body. In comparison with other concurrent methods like FlexProt \cite{shatsky2002flexible} and other rigid body approaches it produced good results in most cases.

\subsection{Geometric hashing method}
Another direction in this field was taken by Shatsky et al. in \cite{shatsky2004method} where they used the \textit{geometric hashing method}. They took 3 atoms or 3 $\alpha$-carbon atom triplets (both can be done) from the protein chains. From n atoms, among $^nC_3$ triplets, they take triplets with some restrictions. At this point they have a set of triplets. Then they prepare a hash table with sides of the triangle as keys to the hash table. Then they selectively align atoms of the two chains using the hash table.

\subsection{ProteinDBS}
ProteinDBS \cite{shyu2004proteindbs} is another method that uses some common features of CBIR (Content Based Image Retrieval) to compare $\alpha$-carbon distance matrix images. This method is much faster than the previous ones as it compares only some specific image features. Notably however, the feature preprocessing done by ProteinDBS is computationally expensive.

\subsection*{TM-align and SP-align}
TM-align \cite{zhang2005tm,zhang2010fast} is one of the most well known protein structure alignment methods. It first finds an optimal alignment and an alignment matrix and then computes the TM-Score as a measure of structural similarity. Another notable method is SP-Align \cite{yang2012new} which employs a similar approach but differs in the alignment algorithm and alignment score.
\begin{figure*}[ht]
\centering
\setlength{\tabcolsep}{1em}
		\begin{tabular}{c c c c} 
		\includegraphics[width=.22\textwidth]{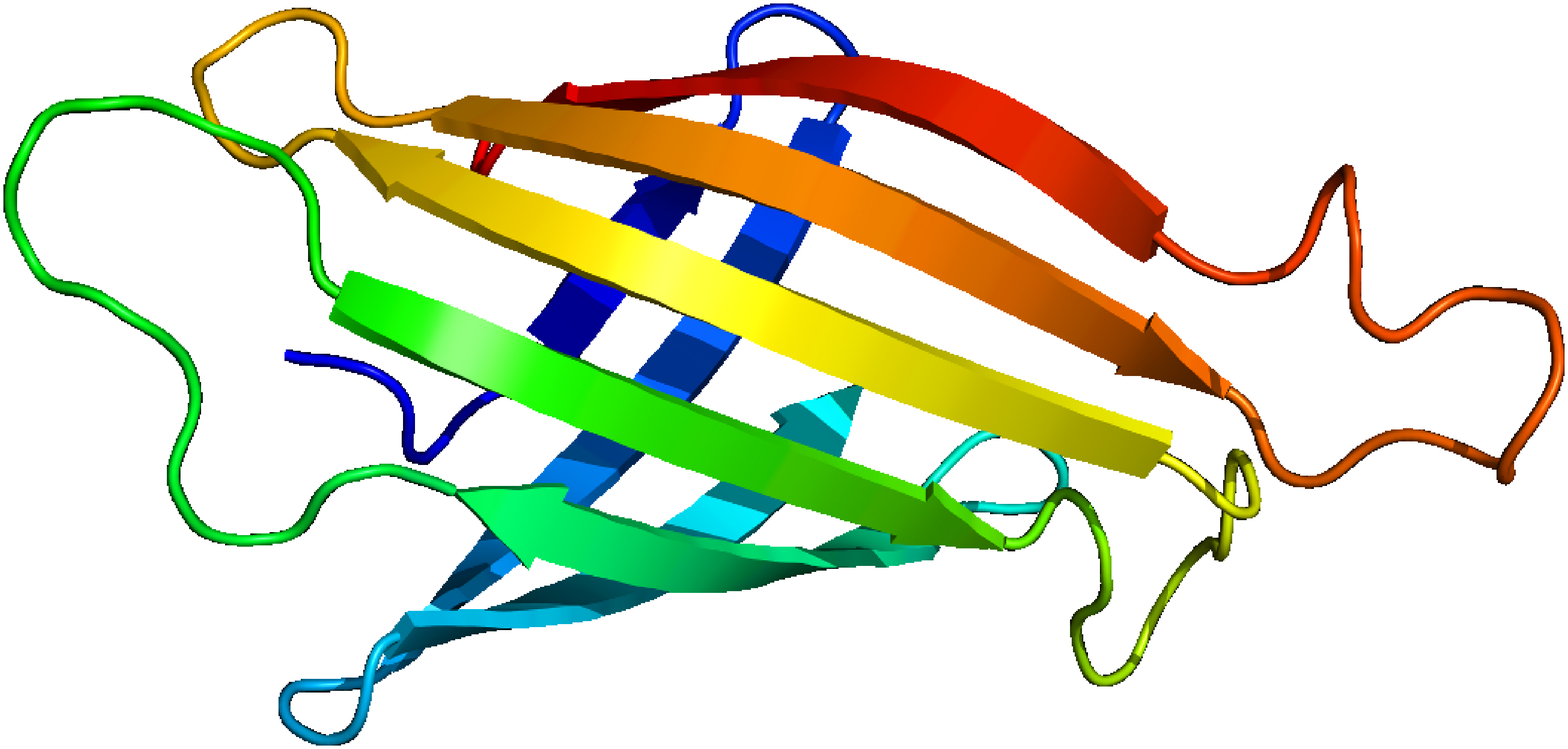}&
		\includegraphics[width=.17\textwidth]{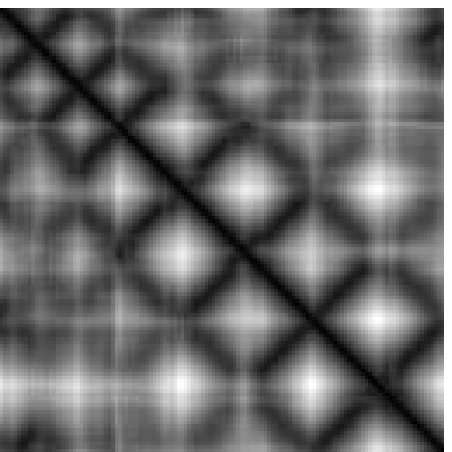}&
		\includegraphics[width=.22\textwidth]{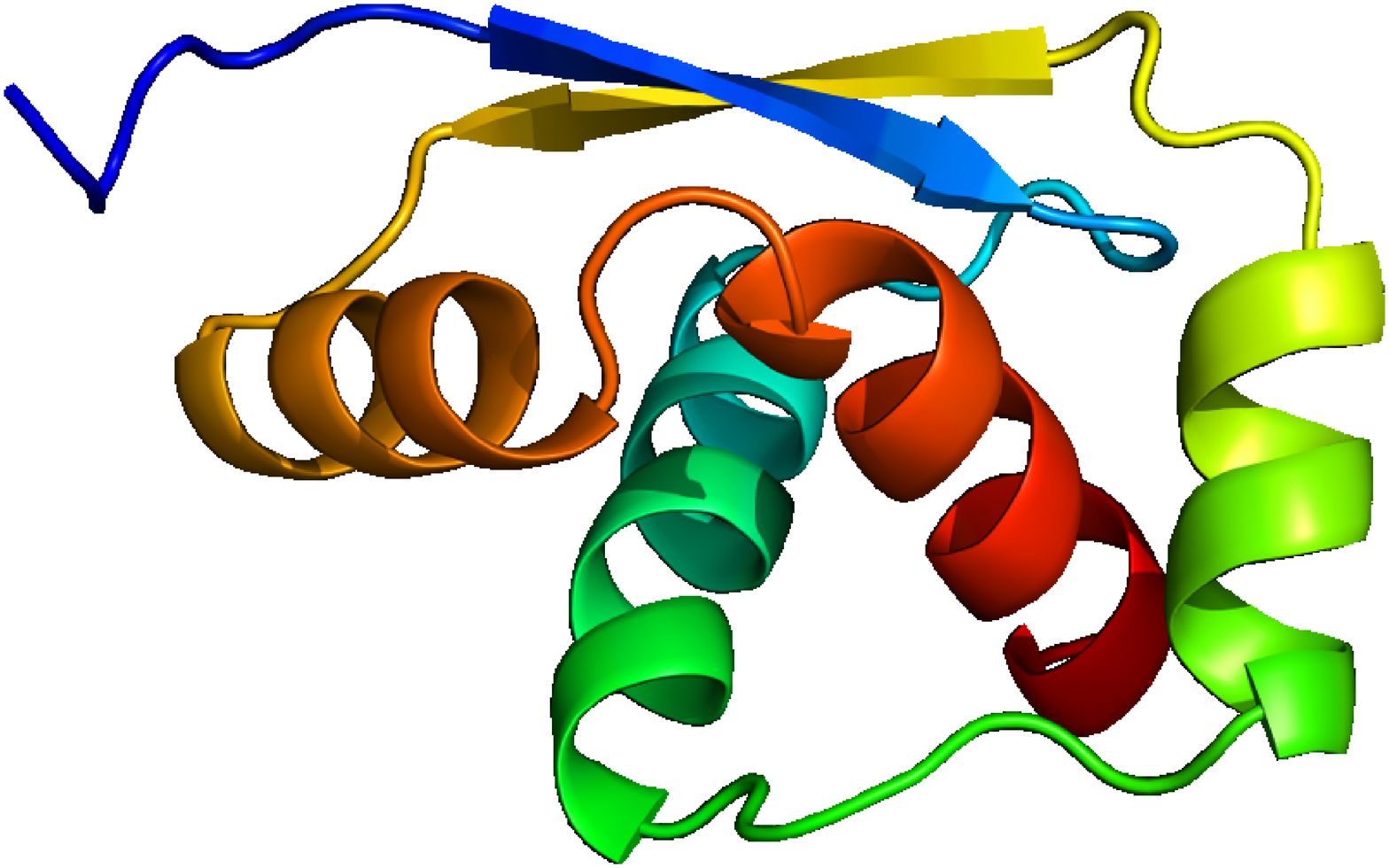}&
		\includegraphics[width=.17\textwidth]{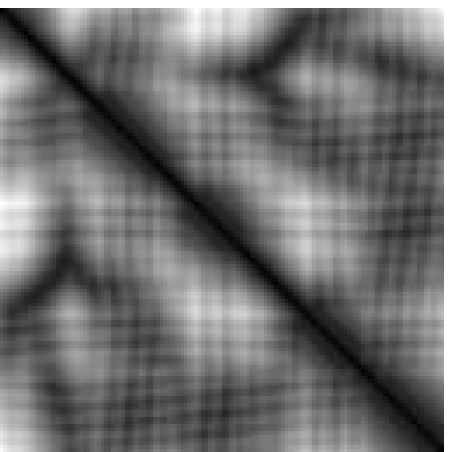}\\
			(a) Domain $d1n4ja\_$ &(b) Extracted feature &(c) Domain $d2efva1$ &(d) Extracted feature\\
		\end{tabular}
	\caption{Original and gray scale images of $\alpha$-carbon distance matrix of 2 proteins $d1n4ja\_$ and $d2efva1$. Representation of $\beta$-sheets are shown in (a), (b) and (c), (d) show the $\alpha$-helices }
	\label{fig:feat_representation}
\end{figure*}
\subsection{Other pattern recognition based approaches}
There exist some other approaches that are based on pattern recognition techniques. These approaches usually involve feature extraction, feature translation and some distance score measures. Perhaps, the most successful feature to this end is the $\alpha$-carbon distance matrix used by Marsolo et al.~\cite{marsolo2006structure}. Marsolo et al. introduced a wavelet based approach that resized the $\alpha$-carbon distance matrices of the protein structures before the actual comparison. In this method the $\alpha$-carbon matrix is considered as a gray scale image and \textit{2D wavelet decomposition} is applied to resize the images to make the feature scale invariant. This method reportedly outperformed existing approaches at that time like DALI and CE, in terms of retrieval accuracy, memory utilization and execution time. This work is one of the prime motivations behind our work.
In the field of pattern recognition, the task of comparing two entity is aided by feature extraction, feature translation and some distance score measure with some distance metric. Among the most successful features, alpha carbon distance matrix is notable. The alpha carbon distance matrix maps three dimensional structure into two dimensional matrix. It is also rotation and translation invariant. There are some methods used alpha carbon distance matrix as feature and their own customized distance metric and distance score calculation algorithm. And there are some methods which extracted some salient features from alpha carbon distance matrix and worked on them.

\subsection{MatAlign}
MatAlign \cite{aung2006matalign} is similar to the method of Marsolo et al. \cite{marsolo2006structure}. It uses alignment of $\alpha$-carbon distance matrix and a score based on the alignment by dynamic programming. MatAlign differs from the structure alignment approaches as it does not align 3D structures directly; rather it aligns two dimensional alpha carbon distance matrix images. Mirceva et al. introduced MASASW \cite{6051424} (Matrix Alignment by Sequence Alignment within Sliding Window) that used Daubechies2 wavelet transform instead of Haar wavelet transform used by Marsolo et al. As the name indicates, MASASW used sliding window to reduce computation. Mirceva et al. have reported that their method outperforms DALI, CE, MatAlign and some other well known methods and also have shown that using Daubechies2 wavelet gives better accuracy than Haar wavelets. Very recently, in our previous work \cite{Karim2015} we have presented a super fast and accurate method to compare protein tertiary structure using an approach based on pattern recognition and computer vision.
\par
In this paper, we attempt to use the same features for structural similarity and compare the performance of our method and statistical significance of distance score of our feature against the most widely used structure alignment based method TM-Align and SP-Align using some well known tools of statistics and machine learning. The method have been previously compared for accuracy and comparison time for protein structure retrieval against MASASW in our previous work \cite{Karim2015}.


\section{Materials and Methods}
\label{secMM}

Our method to protein structure similarity relies upon extraction of features from the three dimensional tertiary structure. We have recently introduced two novel features named \textbf{Co}-occurrence \textbf{M}atrix of the \textbf{O}riented \textbf{Gra}dient of \textbf{D}istance Matrices (CoMOGrad) and Pyramid Histogram of Oriented Gradient (PHOG) in \cite{Karim2015}. These features are extracted from the grayscale images of $\alpha$-carbon distance matrix.

\subsection{Representation of Structures}
From a 3D protein structure we filter only the $\alpha$-carbons. Then we compute their distances from each other and get the $\alpha$-carbon distance matrix. This distance matrix is converted to a grayscale image. The motivation behind this strategy is the observation that the $\alpha$ helix and anti-parallel beta sheets appear as dark lines parallel to the diagonal dark line and parallel beta sheets appear as dark lines normal to the diagonal dark line in that grayscale image. Furthermore, $\beta$-sheets of two strips appear as one dark line normal to the diagonal; $\beta$ sheets of three strips appear as two dark lines normal to the diagonal and one dark line parallel to the diagonal and so on. In general, for a standard $\beta$-sheet, the number of points of co-occurrence of parallel and anti-parallel diagonal lines depends on the number of strips in the $\beta$-sheets. But as different structures have different quantity of $\alpha$-carbons, the matrix dimension would be different. So we scale the distance matrices to the same dimension. With \textit{Bi-cubic Interpolation} we resize each image to the nearest dimension that is a power of 2. Then we apply \textit{Wavelet Transform} to resize all the images to dimension $128\times128$. We then take gradient image of the resized images and extract CoMOGrad and PHOG features from the gradient images as described in the following section. Fig. \ref{fig:feat_representation} shows the tertiary structure of a protein and its gray scale equivalent.

\subsection{CoMOGrad and PHOG}
In our previous work \cite{Karim2015}, we have shown how to extract the novel features CoMOGrad and PHOG. For the sake of completeness here we briefly review the procedure. The gradient angle and magnitude are computed from the gradient of the images mentioned in the previous section. Because of the continuous angle gradient values, we quantize those in \textit{16 bins} for CoMOGrad with bin size \textit{22.5 degrees}. These 16 bins provide us with the CoMOGrad feature vector having 256 features. To extract the \textit{Pyramid histogram of oriented gradient (PHOG)} ~\cite{bosch2013pyramid}, we create a quad tree with the original image taken as the root. Each quad tree node has four children with each child being one fourth of the image corresponding to the parent. We take quad tree up to level 3 which gives us \textit{1+4+4*4+4*4*4=85 nodes}. With \textit{9 bins~(of 40 degrees each)} from the gradient orientation histogram, we get a total of \textit{85*9=768 features} of PHOG. So these 768 PHOG features are added with the 256 features of CoMOGrad  giving us the new novel feature vector of length 1021. The Euclidean distance of two features is used as the measure of dissimilarity of structures being compared.

\subsection{Measurement of CoMOGPhog Score}
Suppose, $f_q$ and $f_i$ denote the feature vectors of the query protein $q$ and a protein $i$ in the database and $N$ is the length of the features. Then the distance score $d_{iq}$ of protein $q$ and $i$ would be calculated according to Equationn~\ref{eqn_dist}.

\begin{eqnarray}\label{eqn_dist}
d_{iq} = \sqrt{\sum_{j=1}^{N} ( |f_q[j]-f_i[j]|)^2}
\end{eqnarray}

Clearly, the above distance measure can be calculated in $O(N)$ time where $N=1021$, the size of our feature vector. As the feature length is independent of the number of alpha carbons, comparison time doesn't vary with the protein length. So, when the features are preprocessed and stored in a database, this method guaranties high scalability. To search nearest structures from a given database, our method needs to compute $d_{iq}$ for each protein $i$ in the database and then sort the results to rank them.
\subsection{Experimental Dataset}
We experimented on 9389 SCOP~\cite{murzin1995scop,fox2014scope} domains (protein structure) and compared our method with SP-Score and TM-Score. In this dataset, there are respectively \textit{624, 1138, 2368} distinct folds, super-families and families available. The list of these domains are listed in the supplementary files.

\section{Results and Discussion}
\label{secRD}
To experimentally verify the statistical significance and performance of structure comparison using the CoMOGPhog score, we have examined a series of well known and widely used statistical measures. These are Posterior probability or P-Value, Mathews Correlation Coefficient (MCC), Receiver Operating Characteristics (ROC), Sensitivity and Specificity. Subsequently, we have compared our results with those of TM-Score and SP-Score. We have taken 9389 protein chains which gives $^{9389}C_2$ pairs. We have computed pairwise score using CoMOGPhog Score, TM-Score and SP-Score. For the latter two, we have used the code provided by the authors in their websites. Implementation of our method is available at \url{http://github.com/rezaulnkarim/CoMOGPhogExtractor}.
 
We are taking the \textit{Structural Classification of Proteins extended (SCOPe)} 2.03 classification \cite{fox2014scope} for the comparison of CoMOGPhog SCore, TM-Score and SP-Score. There are six classifications (Class, Fold, SuperFamily, Family, Protein and Species) which are made from experimentally determined protein structure. We are considering `Family' as its based on sequence meaning proteins in the same family have similarity in their sequence. Though Superfamily (and Fold, Class) classifications are more based on the structure of the protein (as well as our feature), we are still considering Family for an unbiased comparison over the current state of the art methods. 
\begin{figure}[t]
    \centering
    \includegraphics[width=.5\textwidth]{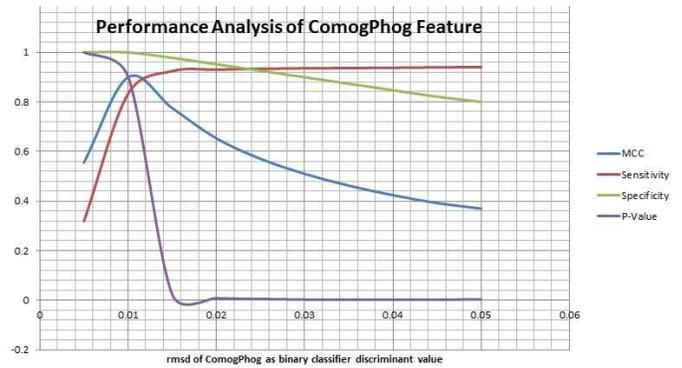}
    \caption{Combined statistics of our method CoMOGPhog score for SuperFamily classification for binary classifier discriminant threshold at score=0.011}
    \label{fig:comogphog_sfamiliy}
\end{figure}
\begin{figure*}[t]
 \centering
\setlength{\tabcolsep}{0em}
 \begin{tabular}{c c c}
   \includegraphics[width=0.35\textwidth]{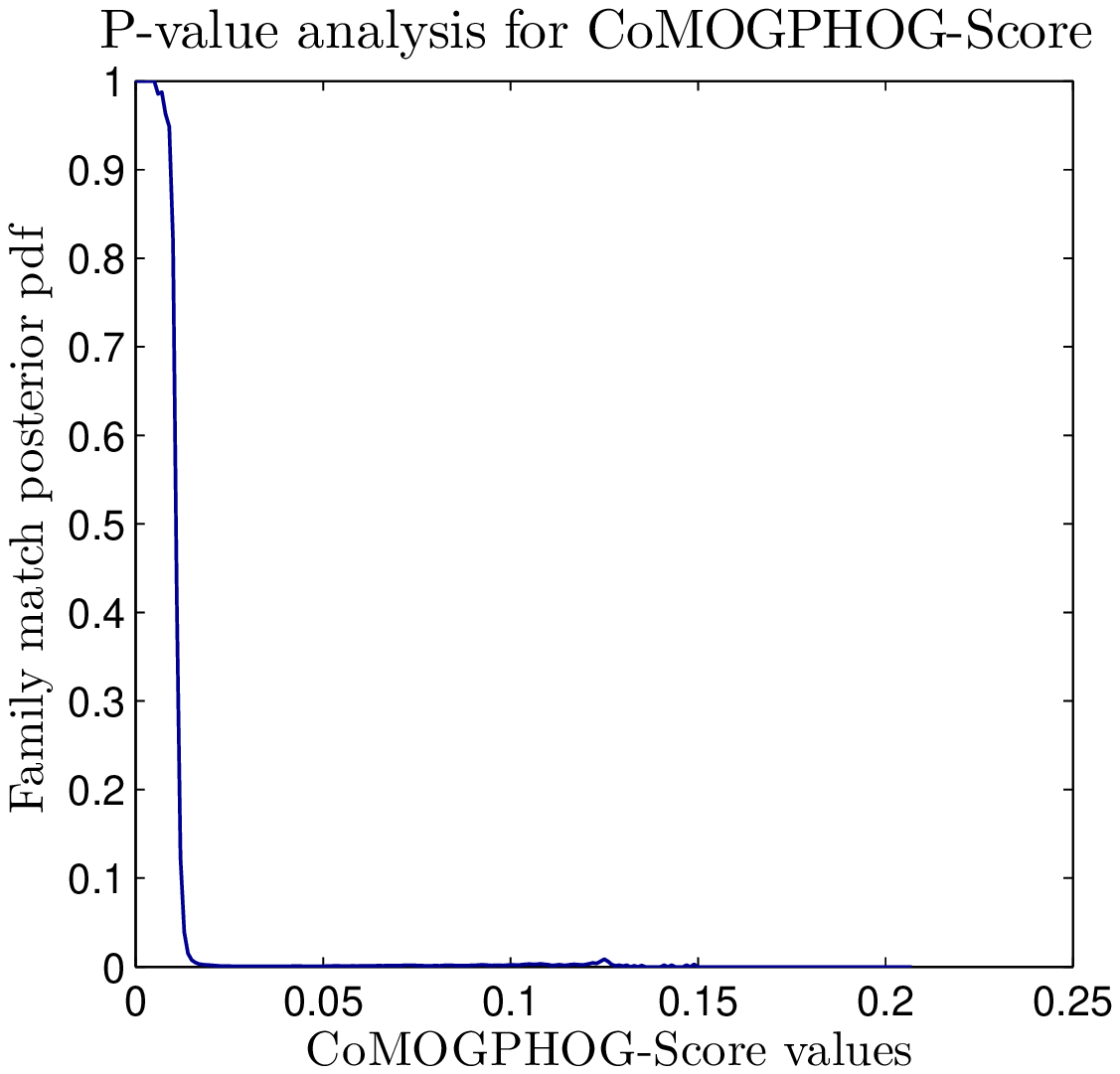}&
   \includegraphics[width=0.35\textwidth]{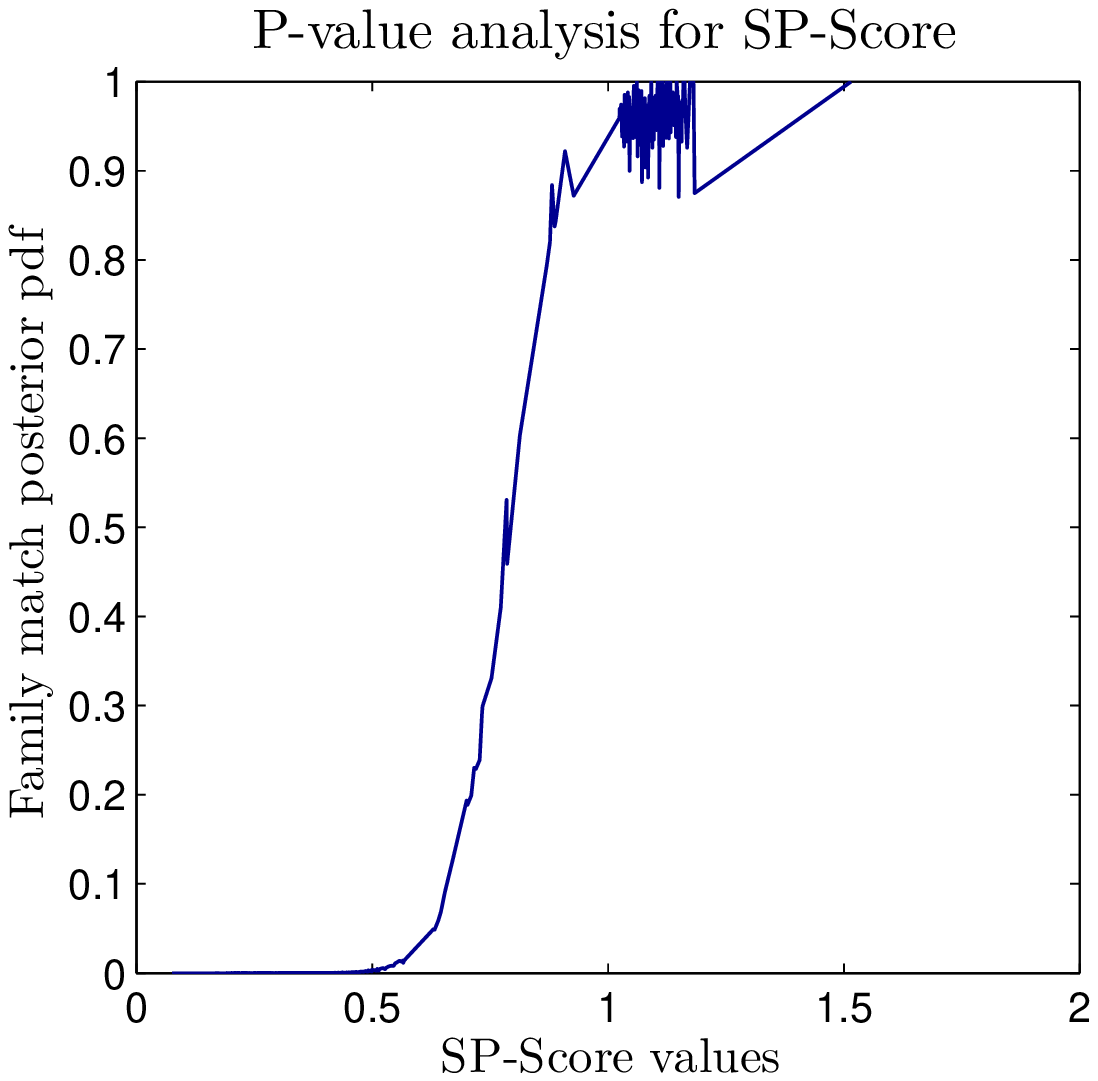}&
  \includegraphics[width=0.35\textwidth]{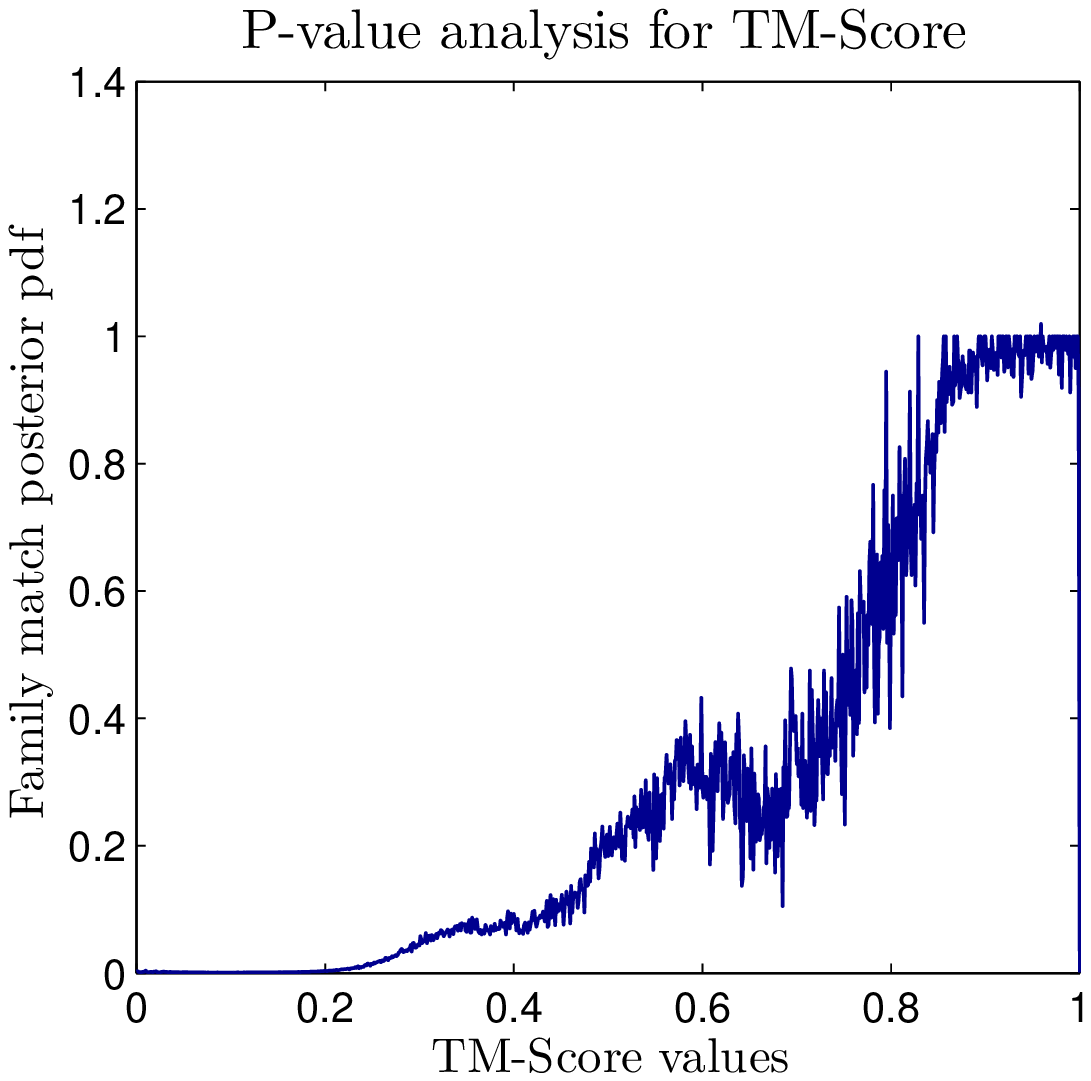}\\ 
   (a)&
  (b)& (c)\\
 \end{tabular}
 \caption{Plot of posterior probability of family match against (a) CoMOGPhog-score (b) SP-Score and (c) TM-Score. \label{fig:PValueAll}}
 \end{figure*}
\subsection{Experimental Results}

\subsubsection{Result for SuperFamily Classification}
Though we did not consider SuperFamily classification for comparing our method with TM-Score and SP-Score, our results for SuperFamily is shown in Figure \ref{fig:comogphog_sfamiliy}. In this experiment, for binary classifier discriminant threshold at score=0.011, MCC value is nearly 0.9, P-value is near 0.84, Sensitivity is near 0.86 and Specificity is near 1 for CoMOGPhog Score. We further use these measurement techniques to compare our method with TM-Score and SP-Score on Family classification.

\subsubsection{Family Match Posterior Probability or P-Value}
\label{P-Value}
Posterior probability or P-Value for a family is defined as the probability of being in the same family for a specific score (CoMOGPhog Score, SP-Score or TM-Score). Using \textit{Bayes Theorem}, we calculate P-Value using Equation~\ref{eqnPValue}.

\begin{footnotesize}
\begin{eqnarray}
\label{eqnPValue}
P(FamilyMatch|score=d)=\frac{P(FamilyMatch, score=d)}{P(score=d)}
\end{eqnarray}
\end{footnotesize}


\begin{figure*}[t]
 \centering
 \setlength{\tabcolsep}{0em}
 \begin{tabular}{c c c}
   \includegraphics[width=0.35\textwidth]{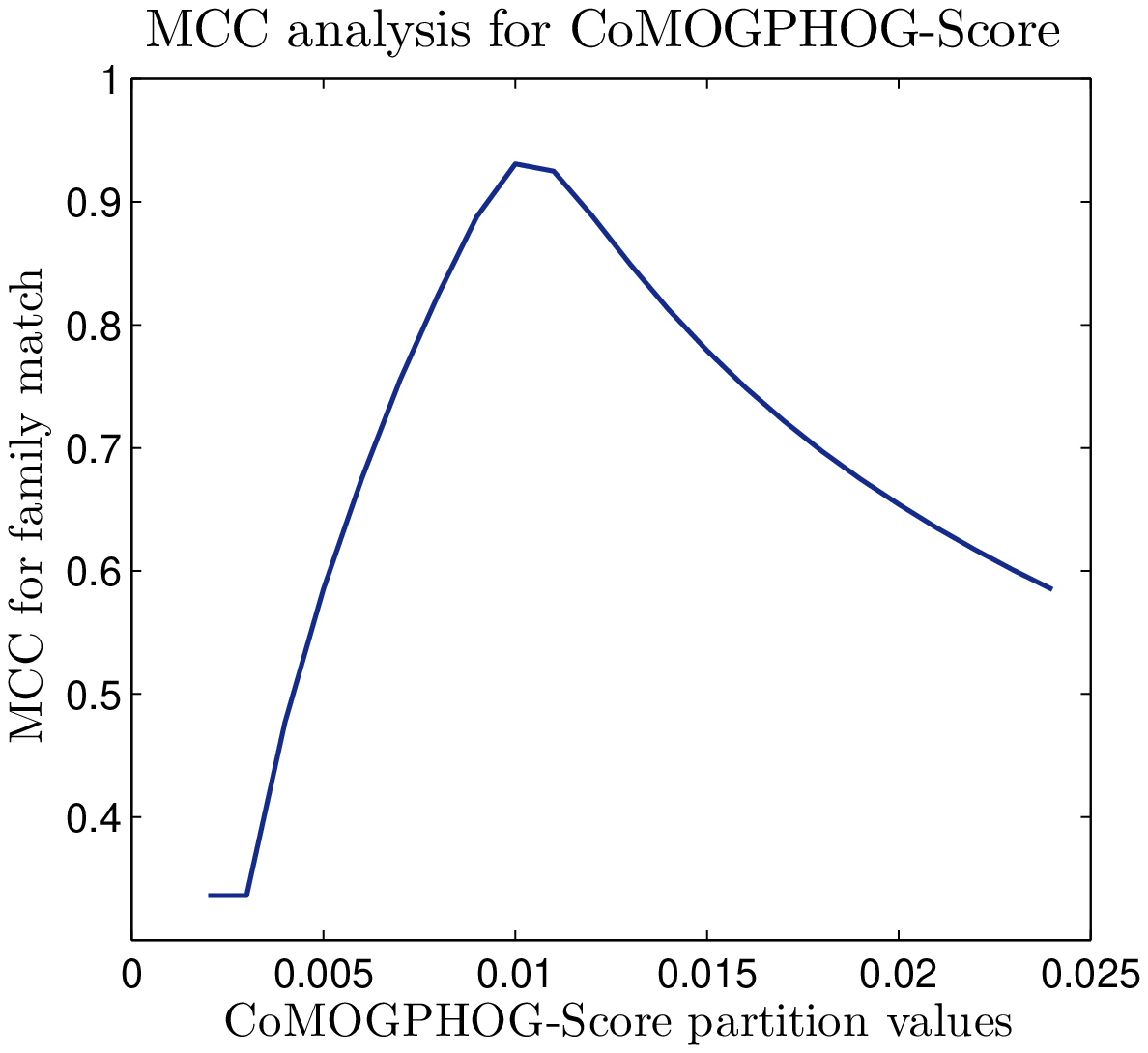}&
   \includegraphics[width=0.35\textwidth]{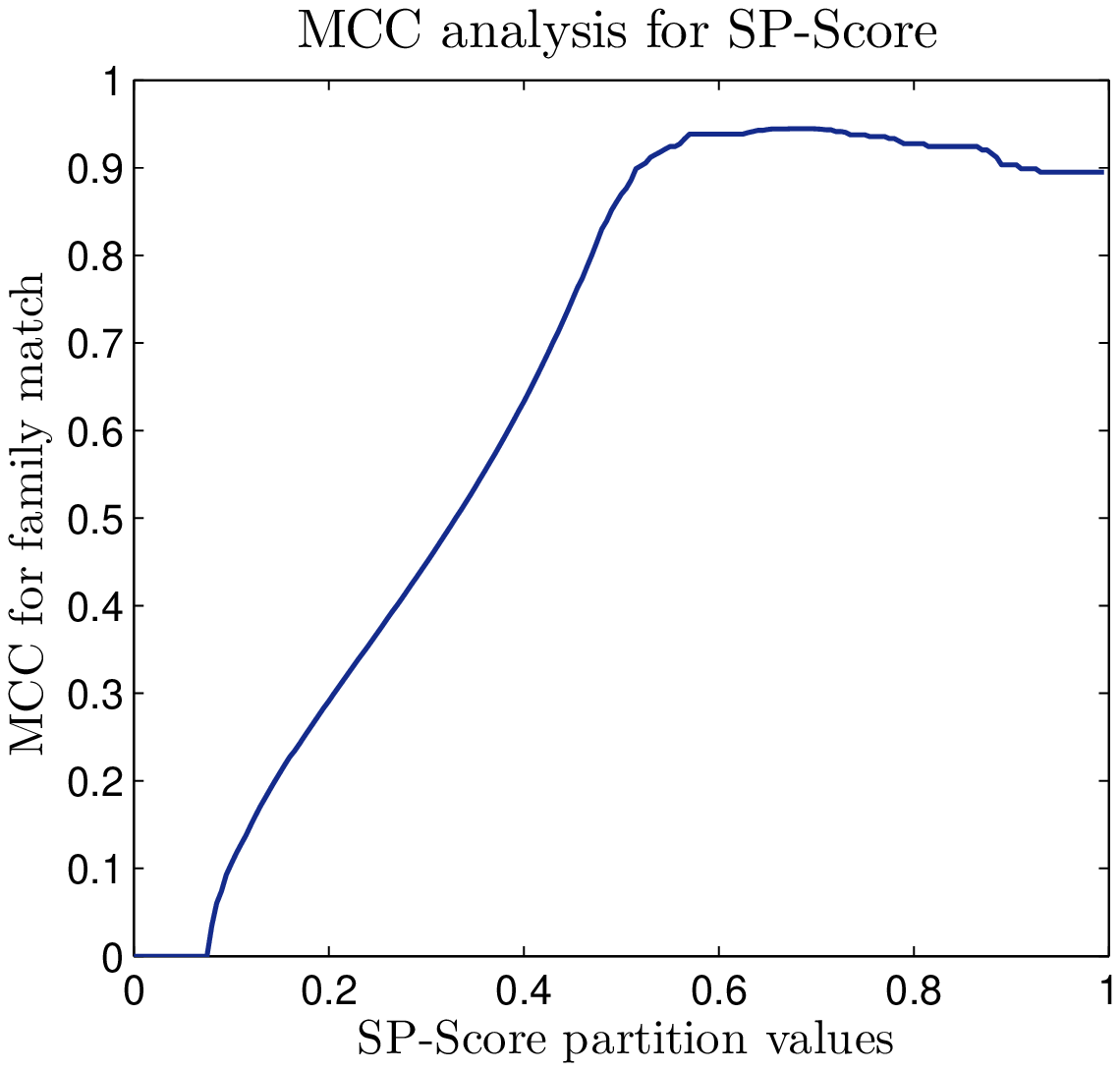}&
   \includegraphics[width=0.35\textwidth]{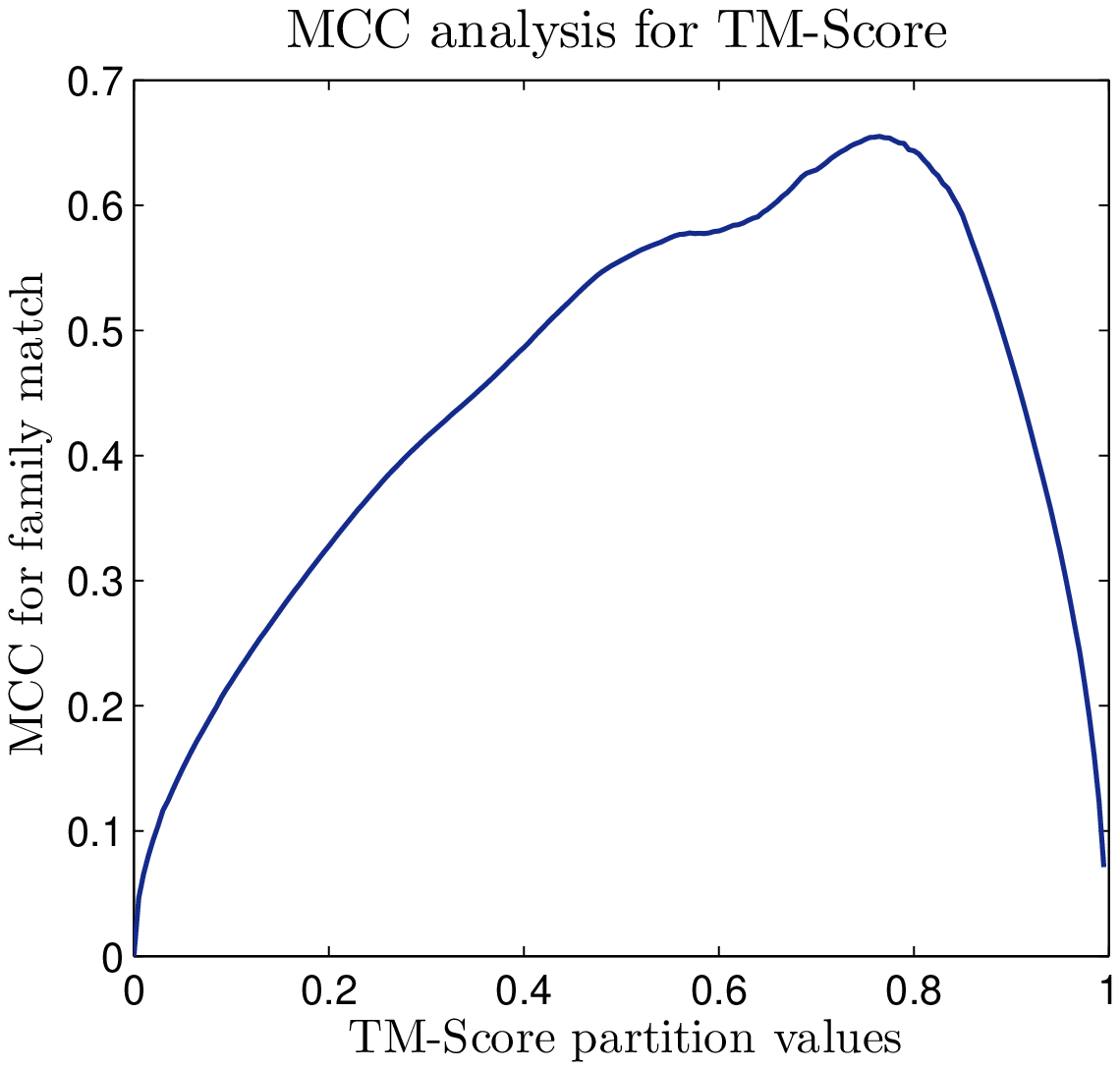}\\
   (a)&  (b)&(c)\\
 \end{tabular}
 \caption{Plot of MCC values against (a) CoMOGPhog-Score (b) SP-Score and (c) TM-Score values. \label{fig:MCC_ALL}}
 \end{figure*}
We have plotted line graph for family match posterior probability for CoMOGPhog Score, SP-Score and TM-Score for all the ($^{9389}C_2$) pairs in  Fig.~\ref{fig:PValueAll}. From the posterior probability or P-Value, the goodness of a distance or score measure can be justified. For a binary classifier discriminant function, in the ideal case, the discriminant function will divide the region into two parts with a vertical line at a specific value of discriminant function, say $d$. Then P-value will be near to 1 below $d$ and near to 0 above $d$ or vice versa. In other words, ideally, the plot should be like a step function, either step down or step up, depending on the definition of the discriminant function.
For SP-Score and TM-Score, the higher the value, the more is the similarity. 
So, the plot of P-Value is expected to be like a step up function. 
On the contrary, for the CoMOGPhog score, the smaller the distance, the more is the similarity in tertiary structure implying more chance of being in the same family. 
So, the plot of P-Value for CoMOGPhog score is expected to be like a step down function. 
From Figure~\ref{fig:PValueAll}(a) it is clearly evident that our score can fix a discriminant distance value $d$ (nearly 0.011) below which P-Value is near to 1 and above which P-Value is near to 0. And the plot is nearly like a step down function. From Figure~\ref{fig:PValueAll}(c), it is clear that for TM-Score, such a point can't precisely be fixed.
And in spite of showing a step up-like trend, the plot of TM-Score is not really a step function as it exhibits too much fluctuations. Similar traits are found in the case of SP-Score (see Figure~\ref{fig:PValueAll}(b)). This clearly indicates the superiority of CoMOGPhog score over SP-Score and TM-Score in terms of family match.


\begin{figure*}[ht]
  \centering
\begin{subfigure}{0.45\textwidth}
\includegraphics[width=\textwidth]{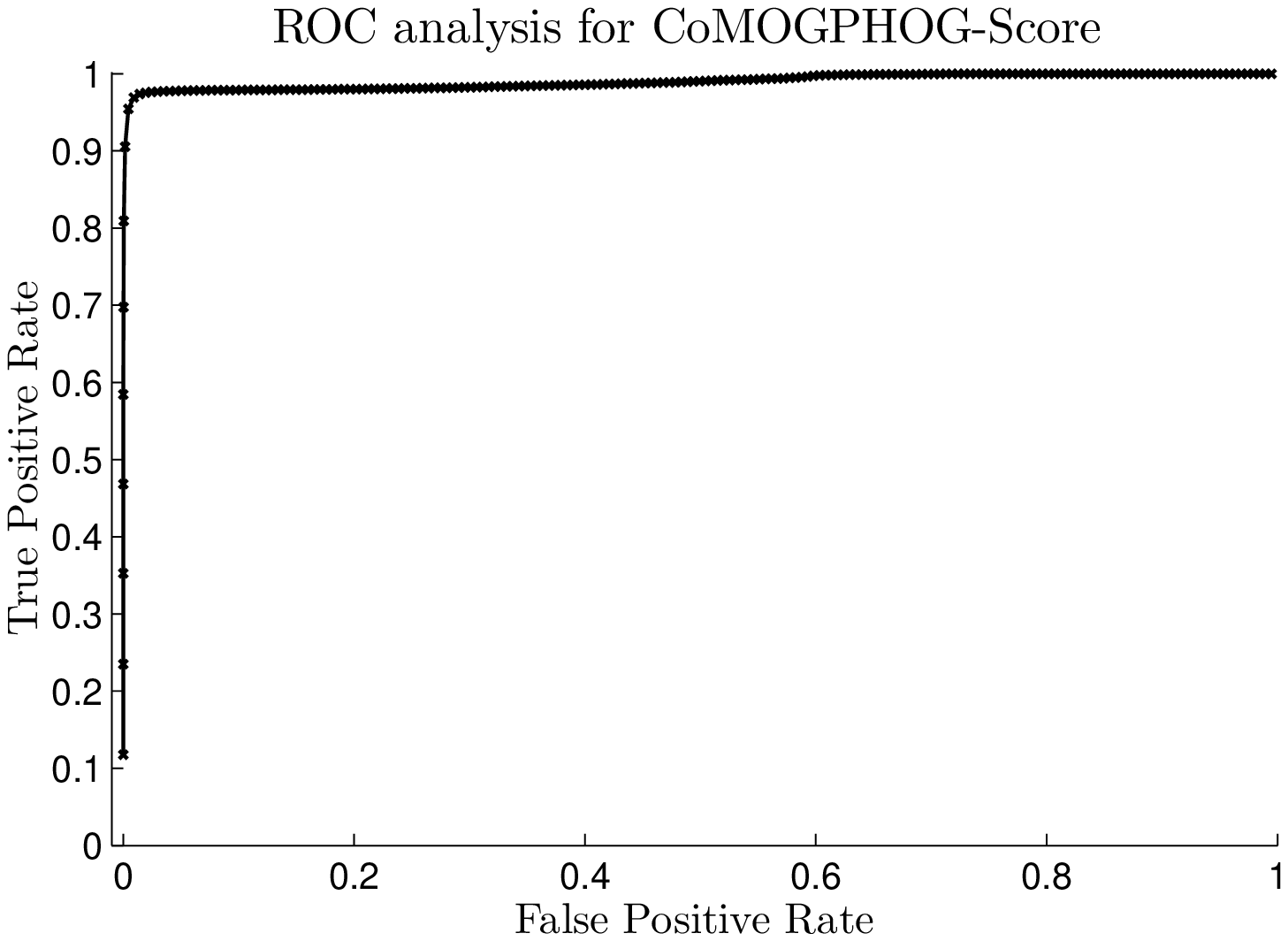}
\caption{CoMOGPhog-Score}
\label{fig:subim1}
\end{subfigure}
 \begin{subfigure}{0.45\textwidth}
\includegraphics[width=\textwidth]{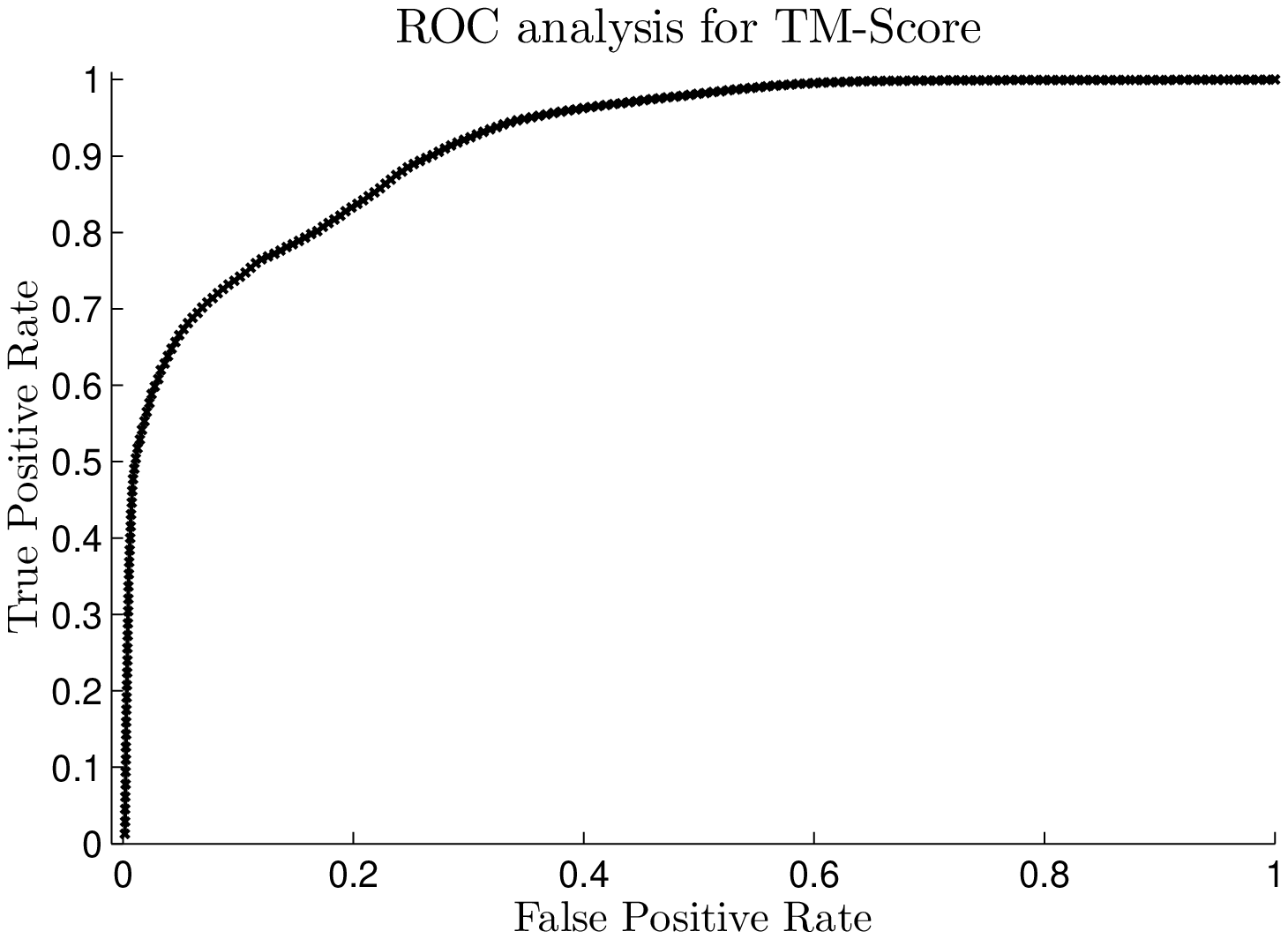}
\caption{TM-Score}
\label{fig:subim3}
\end{subfigure}

\begin{subfigure}{0.45\textwidth}
\includegraphics[width=\textwidth]{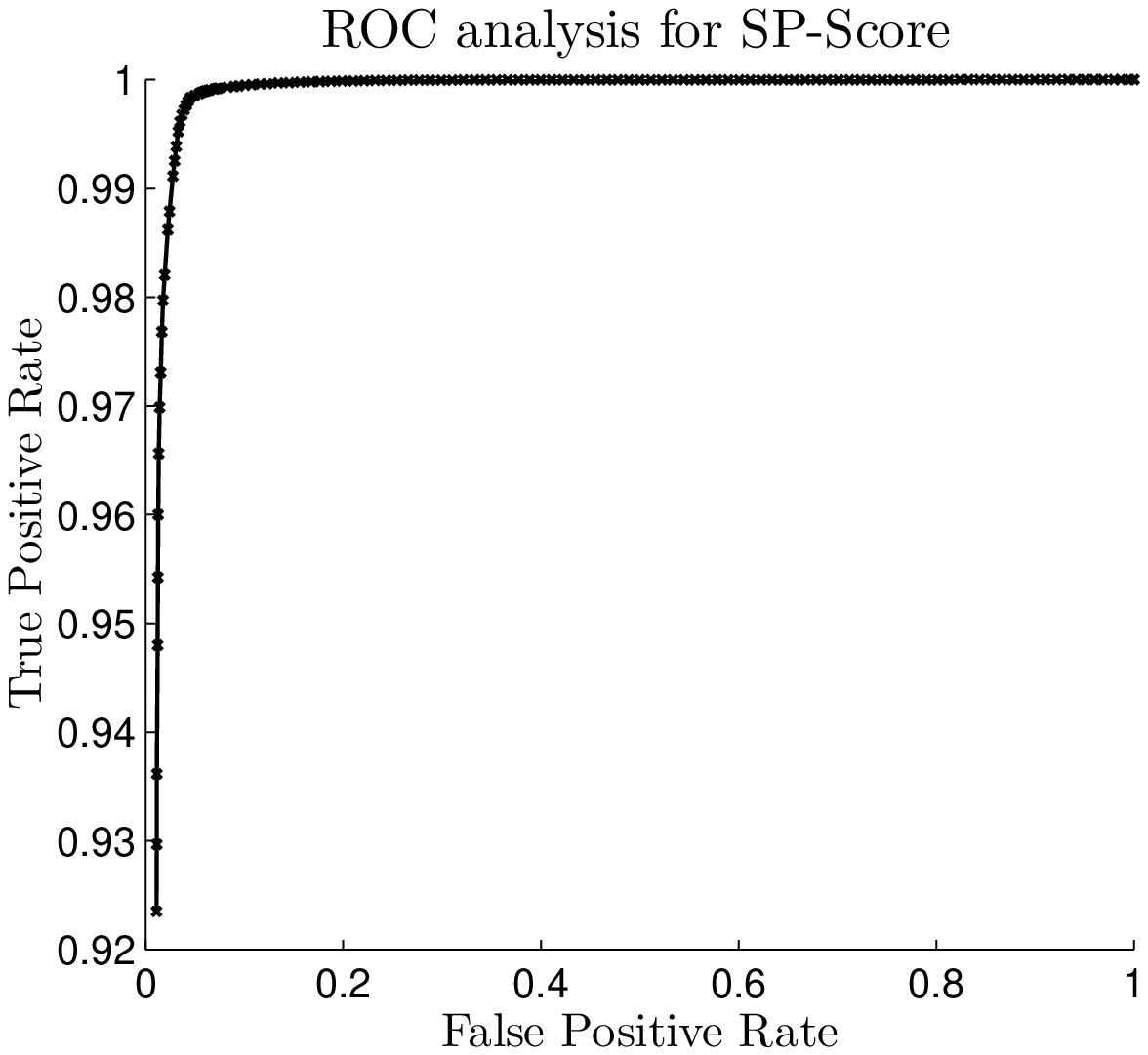}
\caption{SP-Score}
\label{fig:subim2}
\end{subfigure}

\caption{Receiver operating characteristic curve for (a) CoMOGPhog-Score, (b) TM-Score and (b) SP-Score \label{fig:ROI_both}}
\end{figure*}

\subsubsection{Mathews Correlation Coefficient for Binary Classification}
\label{MCC}
{The Matthews Correlation Coefficient (MCC)} is widely used in machine learning as a measure of the quality of binary (two-class) classifications. It takes true-false positives and negatives into account. MCC is generally regarded as a balanced measure which can be used even if the classes are of different sizes. The MCC is in essence a correlation coefficient between the observed and predicted binary classifications. It returns a value between -1 and +1. A coefficient of +1 represents a perfect prediction, 0 indicates that the prediction is no better than random and -1 indicates total disagreement between prediction and observation.
The computation of MCC is done using Equation~\ref{eqnMCC}.
\begin{footnotesize}
\begin{equation}\label{eqnMCC}
MCC=\frac{TP \times TN - FP \times FN}{\sqrt{(TP+FP)\times(TP+FN)\times(TN+FP)\times(TN+FN)}}
\end{equation}
\end{footnotesize}
 

To examine the performance of binary classifiers designed with SP-Score, TM-Score and CoMOGPhog score as discriminants, we have evaluated MCC values with binary classifier discriminant/partition at various values of the scores. For a good classifier, the MCC plot will be a convex plot with peak at the point of best discriminant value. From Fig.~\ref{fig:MCC_ALL}, it is clear that MCC plot for CoMOGPhog score and SP-Score are convex and both have peak values at nearly 0.94 at distance 0.011 for CoMOGPhog score and distance 0.68 for SP-Score. \\
Also Fig.~\ref{fig:MCC_ALL} (c) depicts that MCC plot for TM-Score is convex and has peak value at nearly 0.65 near at TM-Score value = 0.78. This observation suggests the superiority of binary classifier for family with CoMOGPhog score over TM-Score. And it is clear that the MCC plot of CoMOGPhog score is more convex than that of both SP-Score and TM-Score which clearly suggests the superiority of the former.


\subsubsection{Receiver operating characteristic}
\label{ROC}
In statistics, a  receiver operating characteristic (ROC) or an ROC curve is a graphical plot that illustrates the performance of a binary classifier system as its discrimination threshold is varied. The curve is created by plotting the true positive rate (also known as the \emph{sensitivity} in biomedical informatics and \emph{recall} in machine learning) against the false positive rate (specificity, fall-out) at various threshold settings. The ROC curve is thus the sensitivity as a function of fall-out. The more the curve is in the upper left region and the steeper the curve, the better the score function. 
We have plotted ROC curve for the binary classifier systems using SP-Score, TM-Score and CoMOGPhog score as the discrimination threshold is varied. From Fig.~\ref{fig:ROI_both}, it is clearly evident that 
our score is slightly better than both TM-Score and SP-Score in this regard.

\subsubsection{Discussion}
\label{DISC}
From all the statistical measures considered above, it is evident that CoMOGPhog score performs well corresponding to TM-Score and SP-Score with regards to computing protein tertiary structure (dis)similarity. And, more importantly, both TM-Score and SP-Score needs to compute an alignment matrix first to calculate the alignment score. The beauty of CoMOGPhog score is that there is no need to find an alignment first. 
 More specifically, CoMOGPhog score can be simply calculated by computing the \textit{rmsd} of a feature set without even aligning two structures. This speeds up the computation dramatically. Also, since CoMOGPhog score is based on a fixed length feature vector, the comparison time doesn't vary with the protein length.

However there are some parameters used for feature extraction which can be further experimented. The parameter for CoMOgrad is number of bins for gradient angle orientation quantization bin which is used 16 empirically. The parameters for PHOG are depth of quad tree which is used 3 and number of bins for gradient angle orientation quantization bin which is used 9. Both of these values are defined empirically. We expect by fine tuning these parameters with extensive experiments, both the qualitative and quantitative statistical significance and precision of our method can be further improved. We expect to include those fine tuning in our future works.
\section{Conclusion }
\label{secCon}
In this paper, we have presented CoMOGPhog score, which is a novel and computationally efficient scoring scheme for structural classification and pairwise similarity measurement. CoMOGPhog score proivides more significant score than TM-Score and SP-Score in terms of family match of pairs of protein structures and other statistical measurements. Our score is dependent on novel features extracted from protein distance matrices and is inspired from pattern recognition and computer vision. The effectiveness of our method is tested on standard benchmark structures. 


\bibliographystyle{unsrt}  
\bibliography{comog}

\begin{thebibliography}{10}

\bibitem{chothia1986relation}
Cyrus Chothia and Arthur~M Lesk.
\newblock The relation between the divergence of sequence and structure in
  proteins.
\newblock {\em The EMBO journal}, 5(4):823, 1986.

\bibitem{greene2007cath}
Lesley~H Greene, Tony~E Lewis, Sarah Addou, Alison Cuff, Tim Dallman, Mark
  Dibley, Oliver Redfern, Frances Pearl, Rekha Nambudiry, Adam Reid, et~al.
\newblock The cath domain structure database: new protocols and classification
  levels give a more comprehensive resource for exploring evolution.
\newblock {\em Nucleic acids research}, 35(suppl 1):D291--D297, 2007.

\bibitem{mukherjee2009mm}
Srayanta Mukherjee and Yang Zhang.
\newblock Mm-align: a quick algorithm for aligning multiple-chain protein
  complex structures using iterative dynamic programming.
\newblock {\em Nucleic acids research}, 37(11):e83--e83, 2009.

\bibitem{gao2008dbd}
Mu~Gao and Jeffrey Skolnick.
\newblock Dbd-hunter: a knowledge-based method for the prediction of
  dna--protein interactions.
\newblock {\em Nucleic acids research}, 36(12):3978--3992, 2008.

\bibitem{shoichet2012structure}
Brian~K Shoichet and Brian~K Kobilka.
\newblock Structure-based drug screening for g-protein-coupled receptors.
\newblock {\em Trends in pharmacological sciences}, 33(5):268--272, 2012.

\bibitem{kitchen2004docking}
Douglas~B Kitchen, H{\'e}l{\`e}ne Decornez, John~R Furr, and J{\"u}rgen
  Bajorath.
\newblock Docking and scoring in virtual screening for drug discovery: methods
  and applications.
\newblock {\em Nature reviews Drug discovery}, 3(11):935--949, 2004.

\bibitem{hasegawa2009advances}
Hitomi Hasegawa and Liisa Holm.
\newblock Advances and pitfalls of protein structural alignment.
\newblock {\em Current opinion in structural biology}, 19(3):341--348, 2009.

\bibitem{biasini2014swiss}
Marco Biasini, Stefan Bienert, Andrew Waterhouse, Konstantin Arnold, Gabriel
  Studer, Tobias Schmidt, Florian Kiefer, Tiziano~Gallo Cassarino, Martino
  Bertoni, Lorenza Bordoli, et~al.
\newblock Swiss-model: modelling protein tertiary and quaternary structure
  using evolutionary information.
\newblock {\em Nucleic acids research}, page gku340, 2014.

\bibitem{godzik1996structural}
Adam Godzik.
\newblock The structural alignment between two proteins: Is there a unique
  answer?
\newblock {\em Protein science}, 5(7):1325--1338, 1996.

\bibitem{holm1997dali}
Liisa Holm and Chris Sander.
\newblock Dali/{FSSP} classification of three-dimensional protein folds.
\newblock {\em Nucleic acids research}, 25(1):231--234, 1997.

\bibitem{holm1993protein}
Liisa Holm and Chris Sander.
\newblock Protein structure comparison by alignment of distance matrices.
\newblock {\em Journal of molecular biology}, 233(1):123--138, 1993.

\bibitem{shindyalov1998protein}
Ilya~N Shindyalov and Philip~E Bourne.
\newblock Protein structure alignment by incremental combinatorial extension
  (ce) of the optimal path.
\newblock {\em Protein engineering}, 11(9):739--747, 1998.

\bibitem{xu2010significant}
Jinrui Xu and Yang Zhang.
\newblock How significant is a protein structure similarity with tm-score= 0.5?
\newblock {\em Bioinformatics}, 26(7):889--895, 2010.

\bibitem{zhang2005tm}
Yang Zhang and Jeffrey Skolnick.
\newblock Tm-align: a protein structure alignment algorithm based on the
  tm-score.
\newblock {\em Nucleic acids research}, 33(7):2302--2309, 2005.

\bibitem{yang2012new}
Yuedong Yang, Jian Zhan, Huiying Zhao, and Yaoqi Zhou.
\newblock A new size-independent score for pairwise protein structure alignment
  and its application to structure classification and nucleic-acid binding
  prediction.
\newblock {\em Proteins: Structure, Function, and Bioinformatics},
  80(8):2080--2088, 2012.

\bibitem{aung2006matalign}
Zeyar Aung and Kian-Lee Tan.
\newblock {MatAlign}: precise protein structure comparison by matrix alignment.
\newblock {\em Journal of bioinformatics and computational biology},
  4(06):1197--1216, 2006.

\bibitem{6051424}
G.~Mirceva, I.~Cingovska, Z.~Dimov, and D.~Davcev.
\newblock Efficient approaches for retrieving protein tertiary structures.
\newblock {\em Computational Biology and Bioinformatics, IEEE/ACM Transactions
  on}, 9(4):1166--1179, July 2012.

\bibitem{Karim2015}
Rezaul Karim, Mohd Momin~Al Aziz, Swakkhar Shatabda, M.~Sohel Rahman,
  Md~Abul~Kashem Mia, Farhana Zaman, and Salman Rakin.
\newblock Comograd and phog: From computer vision to fast and accurate protein
  tertiary structure retrieval.
\newblock {\em Scientific Reports}, 5:13275 EP --, Aug 2015.
\newblock Article.

\bibitem{orengo1996ssap}
Christine~A Orengo and William~R Taylor.
\newblock Ssap: sequential structure alignment program for protein structure
  comparison.
\newblock {\em Computer methods for macromolecular sequence analysis}, 1996.

\bibitem{ye2003flexible}
Yuzhen Ye and Adam Godzik.
\newblock Flexible structure alignment by chaining aligned fragment pairs
  allowing twists.
\newblock {\em Bioinformatics}, 19(suppl 2):ii246--ii255, 2003.

\bibitem{shatsky2002flexible}
Maxim Shatsky, Ruth Nussinov, and Haim~J Wolfson.
\newblock Flexible protein alignment and hinge detection.
\newblock {\em Proteins: Structure, Function, and Bioinformatics},
  48(2):242--256, 2002.

\bibitem{shatsky2004method}
Maxim Shatsky, Ruth Nussinov, and Haim~J Wolfson.
\newblock A method for simultaneous alignment of multiple protein structures.
\newblock {\em Proteins: Structure, Function, and Bioinformatics},
  56(1):143--156, 2004.

\bibitem{shyu2004proteindbs}
Chi-Ren Shyu, Pin-Hao Chi, Grant Scott, and Dong Xu.
\newblock Proteindbs: a real-time retrieval system for protein structure
  comparison.
\newblock {\em Nucleic Acids Research}, 32(suppl 2):W572--W575, 2004.

\bibitem{zhang2010fast}
Lei Zhang, James Bailey, Arun~S Konagurthu, and Kotagiri Ramamohanarao.
\newblock A fast indexing approach for protein structure comparison.
\newblock {\em BMC bioinformatics}, 11(1):1, 2010.

\bibitem{marsolo2006structure}
Keith Marsolo, Srinivasan Parthasarathy, and Kotagiri Ramamohanarao.
\newblock Structure-based querying of proteins using wavelets.
\newblock In {\em Proceedings of the 15th ACM international conference on
  Information and knowledge management}, pages 24--33. ACM, 2006.

\bibitem{bosch2013pyramid}
A~Bosch and A~Zisserman.
\newblock Pyramid histogram of oriented gradients (phog).
\newblock {\em Univ. Oxford Visual Geometry Group}, 2013.

\bibitem{murzin1995scop}
Alexey~G Murzin, Steven~E Brenner, Tim Hubbard, and Cyrus Chothia.
\newblock Scop: a structural classification of proteins database for the
  investigation of sequences and structures.
\newblock {\em Journal of molecular biology}, 247(4):536--540, 1995.

\bibitem{fox2014scope}
Naomi~K Fox, Steven~E Brenner, and John-Marc Chandonia.
\newblock Scope: Structural classification of proteins?extended, integrating
  scop and astral data and classification of new structures.
\newblock {\em Nucleic acids research}, 42(D1):D304--D309, 2014.

\end{thebibliography}

\end{document}